\documentclass{article}

\usepackage{microtype}
\usepackage{graphicx}
\usepackage{subcaption}
\usepackage{booktabs} % for professional tables
\usepackage{tabularx}
\usepackage{hyperref}

\usepackage[preprint]{icml2026}

\usepackage{amsmath}
\usepackage{amssymb}
\usepackage{mathtools}
\usepackage{amsthm}
\usepackage[table]{xcolor}   % 提供 rowcolor, cellcolor
\usepackage{booktabs}        % \toprule \midrule \bottomrule
\usepackage{multirow}        % 若之后需要多行合并
\usepackage{colortbl}        % 扩展表格颜色支持（可选）
\usepackage{xcolor} % 或者 \usepackage{color}

\usepackage[capitalize,noabbrev]{cleveref}

\theoremstyle{plain}
\newtheorem{theorem}{Theorem}[section]
\newtheorem{proposition}[theorem]{Proposition}

\theoremstyle{definition}

\theoremstyle{remark}

\usepackage[textsize=tiny]{todonotes}

\icmltitlerunning{Submission and Formatting Instructions for ICML 2026}

\begin{document}

\twocolumn[
  \icmltitle{Distribution-Aware Reward Estimation for Test-Time Reinforcement Learning}

  \begin{icmlauthorlist}
    \icmlauthor{Bodong Du}{hkust}
    \icmlauthor{Xuanqi Huang}{hkust}
    \icmlauthor{Xiaomeng Li}{hkust}
  \end{icmlauthorlist}

  \icmlaffiliation{hkust}{Department of Electronic and Computer Engineering, 
  The Hong Kong University of Science and Technology, Hong Kong, China}

  \icmlcorrespondingauthor{Xiaomeng Li}{eexmli@ust.hk}

  \icmlkeywords{Machine Learning, ICML}

  \vskip 0.3in
]



\printAffiliationsAndNotice{}  % no special notice (required even if empty)

\begin{abstract}
Test-time reinforcement learning (TTRL) enables large language models (LLMs) to self-improve on unlabeled inputs, but its effectiveness critically depends on how reward signals are estimated without ground-truth supervision.
Most existing TTRL methods rely on majority voting (MV) over rollouts to produce rewards, implicitly assuming that the majority rollout provides a reliable learning signal.
We show that this assumption is fragile:
MV reduces the rollout distribution into a single outcome, discarding information about non-majority but correct actions, and yields systematically biased reward estimates.
To address this, we propose \textbf{\underline{D}istribution-\underline{A}ware \underline{R}eward \underline{E}stimation (DARE)}, which shifts reward estimation from a single majority outcome to the full empirical rollout distribution.
DARE further augments this distribution-based reward with an exploration bonus and a distribution pruning mechanism for non-majority rollout exploration and reward denoise, yielding a more informative and robust reward estimation. Extensive experiments on challenging reasoning benchmarks show that DARE improves optimization stability and final performance over recent baselines, achieving relative improvements of 25.3\% on challenging AIME 2024 and 5.3\% on AMC.
\end{abstract}

\section{Introduction}

\begin{figure}[t]
    \centering
    \includegraphics[width=\columnwidth]{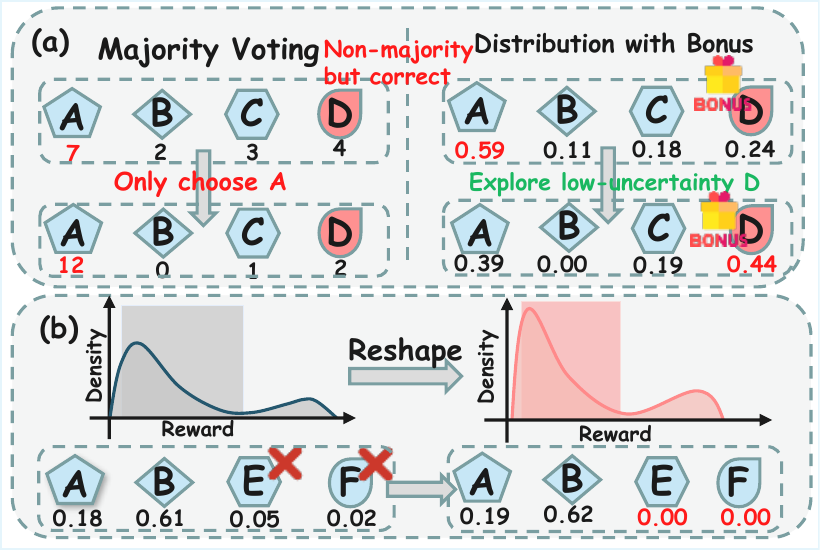}
\caption{\textbf{(a): MV vs. Our Method.} The distribution-based reward with an exploration bonus encourages the model to explore low-uncertainty rollouts and mitigates the confirmation bias of MV. \textbf{(b): Distribution Pruning}  denoise distribution information, reduces reward variance and stabilizes optimization.
}

    \label{fig:introduction}
\end{figure}

Large language models (LLMs) have demonstrated strong capabilities in reasoning and problem solving \cite{ahn2024large,plaat2024reasoning,li2025fundamental,huang2023towards}.
An appealing property of LLMs is their ability to self-improve on unlabeled data via test-time reinforcement learning (TTRL) \cite{zuo2025ttrl},
which enables adaptation at test time without access to external supervision.
In TTRL, the model generates multiple rollouts for a given input and updates its policy using reward signals constructed solely from these self-generated responses.
Because no external labels are available, the quality of these internally constructed rewards plays a central role in determining the effectiveness and stability of test-time optimization.

Most existing TTRL methods 
\cite{zuo2025ttrl,wu2025spinetokenselectivetesttimereinforcement,zhou2025evolvinglanguagemodelslabels} 
construct rewards from multiple rollouts via
\emph{consensus-based} aggregation, such as majority voting (MV)
\cite{chen2024more,majumdar2024generative,gudatiny}.
Concretely, these approaches treat the most frequent answer as a proxy for the optimal action and assign rewards accordingly.  
However, we argue that MV is a suboptimal proxy for reward estimation, as it reduces multiple rollouts to a single majority outcome and discards information carried by non-majority rollouts. In practice, correct responses are not always the most frequent, meaning that useful signals may exist outside the majority \cite{chang2025steplevelverifierguidedhybridtesttime,yu2025pass}. As a result, MV fails to exploit these non-majority but correct signals during optimization.
In addition, our theoretical analysis further shows that this aggregation leads to a systematic mismatch between the MV reward and the expected ground truth reward.
This mismatch lead to confirmation collapse: where early incorrect rewards dominate and are repeatedly reinforced, pushing the model toward suboptimal action space.

% Empirically, this effect becomes more pronounced as rollouts grow increasingly similar, leading to degraded optimization dynamics and a noticeable drop in performance
% (Figure~\ref{fig:introduction}, bottom right).

Based on this analysis, we propose \textbf{Distribution-Aware Reward Estimation (DARE)} for TTRL. 
DARE estimates rewards from an
\textbf{Uncertainty-Aware Empirical Distribution} rather than collapsing outcomes to a single majority vote. 
This distribution-level assignment provides a more reliable guide for policy optimization than MV theoretically. Even when using distribution-based rewards, frequently occurring rollouts can dominate the learning signal, causing less common but potentially correct actions to be underutilized. Importantly, we observed that many of these non-majority but factual correct rollouts tend to exhibit low uncertainty. To leverage this, we introduce an \textbf{Exploration Bonus} that specifically encourages the policy to consider such low-uncertainty, non-majority actions. As illustrated in Figure~\ref{fig:introduction} (a), suppose action $D$ is correct but occurs less frequently (4 times), while action $A$ is more common (7 times). MV simply select $A$, assigning $D$ a zero reward . Our method first assigns $D$ a non-zero reward based on the uncertainty-aware empirical distribution of 0.24 and then boosts it with an exploration bonus. This encourages the policy to gradually increase the reward of $D$, helping the model discover and reinforce less frequent but high-quality actions. And while assigning rewards to all rollouts preserves distributional information, extremely low-quality or noisy responses can propagate through the updates, destabilizing training. For example, in Figure~\ref{fig:introduction} (b), actions $E$ and $F$ have very low empirical probability but can still receive rewards, introducing noise into policy optimization. To mitigate this, we apply \textbf{Distribution Pruning}, removing rollouts with empirical probability below a threshold and renormalizing the remaining rollouts. This process reduces variance in reward signals, stabilizes optimization, and focuses learning on meaningful, high-quality actions.

\noindent Together, distribution-based reward with the exploration bonus and distribution pruning address the key limitations of MV: they reduce bias toward majority but suboptimal actions while filtering out low-quality, noisy rollouts, thereby enabling the policy to learn from both non-majority yet correct actions and reliable high-quality rollouts.

We evaluate DARE on multiple reasoning benchmarks and conduct extensive out-of-distribution (OOD) generalization experiments.
Compared to TTRL, our method consistently improve convergence stability and final performance across multiple reasoning benchmarks and tasks.

In summary, our contributions are:
\begin{itemize}
    \item We identify two key limitations of MV rewards in TTRL: loss of information and a systematical bias lead to confirmation collapse.
    \item We propose DARE, a distribution-aware test-time RL framework that leverages the full rollout distribution for more informative and robust reward estimation.
    \item We show that DARE improves convergence, final performance, and OOD generalization on challenging reasoning benchmarks.
\end{itemize}

\begin{figure*}[t]
    \centering
    \includegraphics[width=\textwidth]{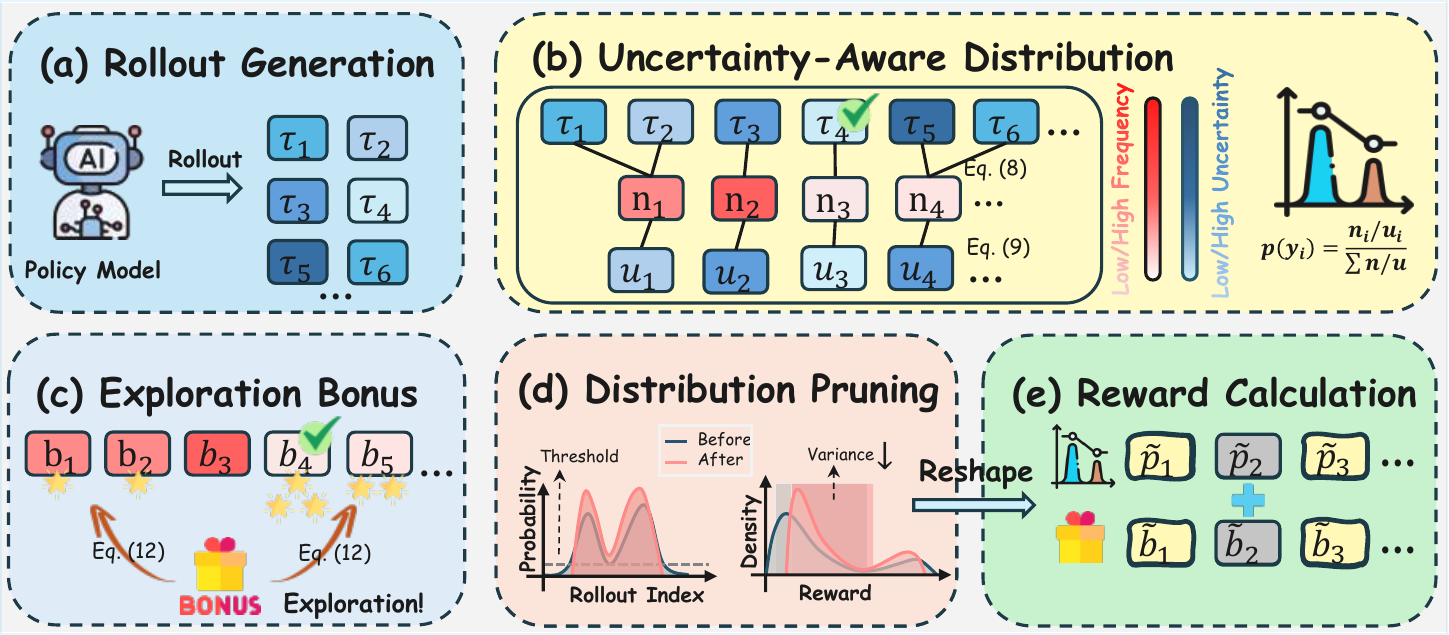}
    \caption{\textbf{Overview of the proposed DARE framework.} 
    Given a test query, multiple rollouts are sampled from the policy model. 
    Rollout-level probabilities are computed based on empirical frequency and uncertainty, followed by exploration bonus and distribution pruning to calculate the final reward used to update policy.}
    \label{fig:pipeline}
\end{figure*}

\section{Theoretical Analysis of Majority Voting as Reward Estimation}

We analyze Majority Voting (MV) as a reward estimator in TTRL.  
Given an input $x$, a policy $\pi_\theta$ generates $M$ rollouts 
$P=\{y_1,\dots,y_M\}$ with empirical distribution $\hat{p}(y)$.  
MV first estimates a pseudo label
\begin{equation}
\hat{y}(P) = \arg\max_y \hat{p}(y),
\end{equation}
and assigns each rollout a rule-based reward \cite{guo2025deepseek}. 
Throughout this section, rewards are treated as random variables and denoted by uppercase $R(\cdot)$.
\begin{equation}
R_{\mathrm{MV}}(y_i; P)
=
R\!\left(y_i,\; \hat{y}(P)\right),
\end{equation}

\subsection{Limitation 1: Loss of Reward-Relevant Information}

MV maps a distribution over samples to a binary labeling induced by the empirical mode, discarding all distributional structure.  
Let $Y \sim p(y)$ denote a random rollout.

\begin{theorem}[Information Collapse under Majority Voting]
The reward signal induced by MV satisfies
\begin{equation}
I(R(Y); R_{\mathrm{MV}}(Y;P)) \;\le\; I(R(Y); Y),
\end{equation}
with strict inequality whenever multiple outputs with distinct rewards have nonzero probability mass.
\end{theorem}

Thus, MV is an information-losing estimator: it compresses the rollout population into a single outcome and eliminates uncertainty and diversity relevant for policy update. This reduction is not merely a theoretical concern, it has concrete consequences in practice. By reducing potentially dozens of rollouts to a single consensus answer, MV discards all non-majority paths, regardless of their intrinsic quality and information. Crucially, correct answers are not always the most frequent in rollouts, especially on complex reasoning tasks where models may generate diverse valid solutions or where the majority reflects a systematic but incorrect bias

\subsection{Limitation 2: Bias under Correlated Rollouts}

We model rollout dependence by an exchangeable latent variable $Z$, such that each rollout is conditionally sampled from $p(y \mid Z)$. The target objective is the marginal expected reward $\mathbb{E}_{y \sim p(y)}[R(y)]$, whereas MV estimates it through the conditional mode induced by $Z$.

\begin{theorem}[Latent-Conditioned Bias of MV]
Assume binary rewards and positively correlated, exchangeable rollouts generated under a latent-variable model.  
Then the rollout-level MV reward satisfies
\begin{equation}
\mathbb{E}\!\left[ R_{\mathrm{MV}}(Y;P) \right]
\;\neq\;
\mathbb{E}_{y \sim p(y)}[R(y)],
\end{equation}
whenever $p(y \mid Z) \neq p(y)$ with positive probability.
\end{theorem}

Hence, MV estimates rewards with respect to a \emph{latent-conditional mode} rather than the marginal expected reward, resulting in a systematic bias in reward estimation.
When such biased rewards are used for policy updates, they naturally induce a self-reinforcing optimization dynamic consistent with confirmation bias.

Together, these results show that MV is neither information-preserving nor unbiased as a reward estimator in TTRL, motivating distribution-aware alternatives that operate directly on the rollout space.

\subsection{Distribution-Based Reward as Marginal Estimation}

We now consider distribution-based reward assignment as a proxy for marginal reward estimation, under the standard assumption that higher-probability rollouts are more likely to be correct, as commonly adopted in self-training frameworks. Each rollout $\hat{y}_m$ is assigned utility as a monotonic transformation of its empirical frequency,
\begin{equation}
R_{\mathrm{dist}}(\hat{y}_m) = g\!\left(\hat{p}(\hat{y}_m)\right),
\end{equation}
where $g(\cdot)$ is a monotonic shaping function.

\begin{proposition}[Marginal Consistency under Exchangeable Rollouts]
\label{prop:marginal_proxy}
Under the latent-variable model, the distribution-based reward satisfies
\begin{equation}
\mathbb{E}\!\left[ R_{\mathrm{dist}}(Y) \right]
=
\mathbb{E}_{y \sim p(y)} \!\left[ g\!\left(p(y)\right) \right],
\end{equation}
where $p(y) = \mathbb{E}_{Z}[p(y\mid Z)]$ is the marginal rollout distribution.  
Accordingly, $R_{\mathrm{dist}}$ assigns utility aligned with the marginal rollout probability in expectation, and provides a policy-consistent proxy signal.
\end{proposition}

\noindent
Thus, unlike MV which estimates a latent-conditional mode, distribution-based reward preserves marginal probability information across modes and avoids conditional bias.  
Motivated by this principle, we develop \textbf{Distribution-Aware Reward Estimation} with  exploration control and distribution pruning.

\section{Method}

We present Distribution-Aware Reward Estimation (DARE), a framework for estimating test-time reward signals that account for both uncertainty and distributional structure in model rollouts. By operating on the empirical rollout population, DARE preserves the full spectrum of reasoning paths, capturing both prevalence and internal uncertainty. This enables exploitation of confident common answers while exploring rare, high-quality trajectories, providing richer feedback at test time.
Figure~\ref{fig:pipeline} illustrates the overall workflow, with steps (a) through (e) corresponding to each key component.

\subsection{Rollout Sampling and Uncertainty-Aware Distribution}

The first step of DARE (Figure~\ref{fig:pipeline}a) is to sample diverse reasoning trajectories in order to construct an empirical distribution over candidate answers.  
Given a test query $q$, the language model generates $M$ rollouts:
\begin{equation}
\{\tau_1, \dots, \tau_M\} \sim \pi_\theta(\cdot \mid q),
\end{equation}
where each rollout $\tau_i$ produces a final answer $y_i, \; i = 1, \dots, M$.  
These rollouts can provide a natural foundation for distribution-aware reward estimation.

To quantify how answers are distributed across rollouts, we measure the prevalence of each candidate answer $\hat{y}$ using empirical frequency:
\begin{equation}
n(\hat{y}) = \sum_{k=1}^{M} \mathbf{1}[\hat{y}_k = \hat{y}],
\end{equation}
which reflects how often a particular outcome is repeatedly generated.  
However, frequency alone is insufficient due to the rollouts' internal quality. To address this, we define a trace-level uncertainty score for each candidate answer as the average token entropy across all rollouts:
\begin{equation}
u(\hat{y}) = \frac{1}{n(\hat{y})} \sum_{k:\hat{y}_k=\hat{y}} \frac{1}{|\tau_k|} \sum_{i \in \tau_k} \sum_j -P_i(j)\log P_i(j),
\end{equation}
where $P_i(j)$ denotes the predicted probability of token $j$-th vocabulary token at position $i$.  
This measure captures the internal consistency of the reasoning process.

By combining prevalence and uncertainty, we define an uncertainty-aware empirical distribution over candidate answers (Figure~\ref{fig:pipeline}b):
\begin{equation}
\hat{p}(\hat{y}) = \frac{n(\hat{y}) / (u(\hat{y})+\epsilon)}{\sum_{\hat{y}'} n(\hat{y}') / (u(\hat{y}')+\epsilon)},
\end{equation}
where $\epsilon > 0$ ensures numerical stability.  
This distribution preserves the overall observed outcomes while reducing bias toward frequent but unreliable rollouts, thereby providing a more faithful estimate of answer quality.

\subsection{Distribution-based Reward and Exploration Bonus}

Based on the uncertainty-aware distribution, we assign a base reward to each rollout according to its final answer:
\begin{equation}
r_{\text{dis}}(y_i) = \hat{p}(y_i),
\end{equation}
which naturally encourages exploitation of frequent and internally consistent responses.  
Nevertheless, even distribution-based rewards can remain biased toward dominant modes, especially when correct but alternative reasoning paths appear infrequently.

To explicitly counteract this effect, we introduce an exploration bonus (Figure~\ref{fig:pipeline}c):
\begin{equation}
b(y_i) = \Big(1 - \frac{n(y_i)}{M}\Big) \cdot \big(1 - u(y_i)\big).
\end{equation}
This bonus assigns additional reward to rollouts that are less frequent with low uncertainty, thereby promoting exploration of promising but non-majority trajectories. Importantly, the uncertainty term prevents amplification of rare but noisy rollouts, ensuring that exploration remains reliable.

The final reward for each rollout is obtained by combining the distribution-based component with the exploration bonus:
\begin{equation}
r(y_i) = r_{\text{dis}}(y_i) + \alpha \, b(y_i),
\end{equation}
where $\alpha \in [0,1]$ controls the strength of exploration.  
This probability-shaped reward encourages diversity without sacrificing stability, mitigating premature collapse to latent-specific modes.

\subsection{Distribution Support Pruning}

Despite probability shaping and exploration incentives, extremely low-probability rollouts may still introduce noise and destabilize optimization.  
DARE therefore performs distribution support pruning by removing rollouts whose empirical probability falls below a threshold $\tau$, followed by renormalization over the retained support (Figure~\ref{fig:pipeline}d):
\begin{equation}
 \tilde{p}(y_i) 
= \frac{\hat{p}(y_i)\,\mathbf{1}[\hat{p}(y_i) \ge \tau]}
       {\sum_{k=1}^{M} \hat{p}(y_k)\,\mathbf{1}[\hat{p}(y_k) \ge \tau]} .
\end{equation}

After pruning, all distribution-dependent statistics are recomputed on the surviving rollouts.  
In particular, $\tilde{n}(y_i)$, $\tilde{u}(y_i)$, and $\tilde{b}(y_i)$ are evaluated using only the retained support.  
The final reshaped reward (Figure~\ref{fig:pipeline}e) is then defined as
\begin{equation}
r(y_i) = \tilde{r}_{\text{dis}}(y_i) + \alpha\, \tilde{b}(y_i)
= \tilde{p}(y_i) + \alpha\, \tilde{b}(y_i) .
\end{equation}
This pruning step removes degenerate low-quality rollouts, reduces reward variance, and mitigates noisy gradient updates, leading to more stable and robust optimization.

\subsection{Test-Time Policy Optimization}

Finally, the refined rollout-level rewards are used to update the policy via GRPO at test time.  
By jointly leveraging uncertainty-aware reward estimation, exploration bonus, and distribution pruning, DARE enables the policy to exploit high confident majority responses while systematically exploring valuable non-majority rollouts, so it can effective mitigate confirmation collapse and result in stable and effective test-time adaptation.

\begin{table*}[t]
\centering

\vspace{-0.8em}
\label{tab:main_results}
\resizebox{\textwidth}{!}{%
\begin{tabular}{@{}c|lcccccc@{}}
\toprule
 & \multirow{2}{*}{\textbf{Method}} 
 & \multicolumn{1}{c}{\textbf{General}} 
 & \multicolumn{3}{c}{\textbf{Mathematical Reasoning}} 
 & \multicolumn{1}{c}{\textbf{Sci. Reasoning}} 
 & \multirow{2}{*}{\textbf{Avg}} \\
\cmidrule(lr){3-3} \cmidrule(lr){4-6} \cmidrule(lr){7-7}
 &  & MMLU-Pro & MATH-500 & AIME 2024 & AMC & GPQA &  \\ 
\midrule

% ===================== Model 1 =====================
\multirow{12}{*}{\rotatebox{90}{\textbf{Qwen2.5-Math-1.5B}}} &
\multicolumn{7}{c}{\textbf{\textit{Prompt methods (training-free)}}} \\

 & Raw model  \cite{yang2024qwen2}
 & 25.8 & 32.7 & 7.7 & 28.6 & 27.8 & 24.5 \\

 & CoT \cite{COT}
 & 27.5 & 34.1 & 8.5 & 29.8 & 28.6 & 25.7 \\

\cmidrule{2-8}
 & \multicolumn{7}{c}{\textbf{\textit{RL methods on training dataset}}} \\

 & GRPO \citep{shao2024deepseekmath} 
 & 33.5 & 72.8 & 14.3 & 46.5 & 26.1 & 42.3 \\

 & REINFORCE \cite{Williams1992SimpleSG} 
 & 32.5 & 72.0 & 14.6 & 46.0 & 25.2 & 40.1 \\

 & REINFORCE++ \cite{REINFORCE+}
 & \underline{34.3} & \underline{72.8} & 15.2 & 46.8 & 25.9 & 41.0 \\

\cmidrule{2-8}
 & \multicolumn{7}{c}{\textbf{\textit{Test-time scaling methods}}} \\

 &  INTUITOR \cite{intuitor}  
 & 31.1 & 70.0 & 12.0 & 44.5 & 23.5 & 38.2 \\

 &  RLPR \cite{RLPR}  
 & 33.9 & \underline{73.2} & \underline{16.0} & 47.0 & \underline{26.5} & 41.3 \\

 &  CO-REWARDING-I \cite{coreward}  
 & 33.1 & 72.5 & 15.5 & 46.5 & 25.8 & 40.7 \\

 &  TTRL \cite{zuo2025ttrl}  
 & \underline{35.6} & 73.0 & 15.8 & \underline{47.3} & 26.1 & \underline{41.5} \\

\rowcolor{blue!10}
 & \textbf{ DARE (Ours)} 
 & \textbf{38.9} 
 & \textbf{73.6} 
 & \textbf{19.8} 
 & \textbf{50.2} 
 & \textbf{28.5} 
 & \textbf{44.2} \\

\midrule
\midrule

% ===================== Model 2 =====================
\multirow{12}{*}{\rotatebox{90}{\textbf{Qwen3-1.7B}}} &
\multicolumn{7}{c}{\textbf{\textit{Prompt methods (training-free)}}} \\

 & Raw model \cite{yang2025qwen3}
 & 36.8 & 55.8 & 11.4 & 32.6 & 26.8 & 32.6 \\

 & CoT \cite{COT}
 & 38.5 & 57.2 & 12.3 & 33.8 & 27.6 & 33.8 \\

\cmidrule{2-8}
& \multicolumn{7}{c}{\textbf{\textit{RL methods on training dataset}}} \\

 & GRPO \citep{shao2024deepseekmath}
 & 44.5 & 77.3 & 24.1 & 53.2 & 31.1 & 48.2 \\

 & REINFORCE \cite{Williams1992SimpleSG}
 & 44.2 & 77.5 & 23.2 & 52.3 & 30.2 & 47.4 \\

 & REINFORCE++ \cite{REINFORCE+}
 & 45.4 & 78.0 & 23.9 & 52.8 & 30.8 & 48.1 \\

\cmidrule{2-8}

 & \multicolumn{7}{c}{\textbf{\textit{Test-time scaling methods}}} \\

 &  INTUITOR \cite{intuitor}  
 & 42.1 & 77.5 & 20.5 & 50.5 & 28.5 & 45.8 \\

 & RLPR \cite{RLPR}  
 & 45.7 & 75.5 & \underline{24.5} & \underline{53.0} & 31.0 & \underline{48.9} \\

 &  CO-REWARDING-I \cite{coreward}  
 & 43.2 & \underline{78.9} & 23.8 & 52.5 & 30.5 & 47.7 \\

 &  TTRL \cite{zuo2025ttrl}  
 & \underline{46.9} & 78.2 & 24.0 & 52.9 & \underline{31.2} & 48.6 \\

\rowcolor{blue!10}
 & \textbf{ DARE (Ours)} 
 & \textbf{48.8} 
 & \textbf{79.6} 
 & \textbf{26.3} 
 & \textbf{55.7} 
 & \textbf{32.7} 
 & \textbf{50.6} \\

\bottomrule
\end{tabular}%
}
\caption{Performance Comparison across two backbones, evaluated on five benchmarks from three task categories.
The best and second best results are \textbf{highlighted} and \underline{underlined}, respectively.}
\vspace{-1em}
\label{tab:exp}
\end{table*}

% -------------------------------
% 第一张图：Qwen2.5-Math-1.5B
% -------------------------------
\begin{figure*}[t]
    \centering
    \includegraphics[width=\textwidth]{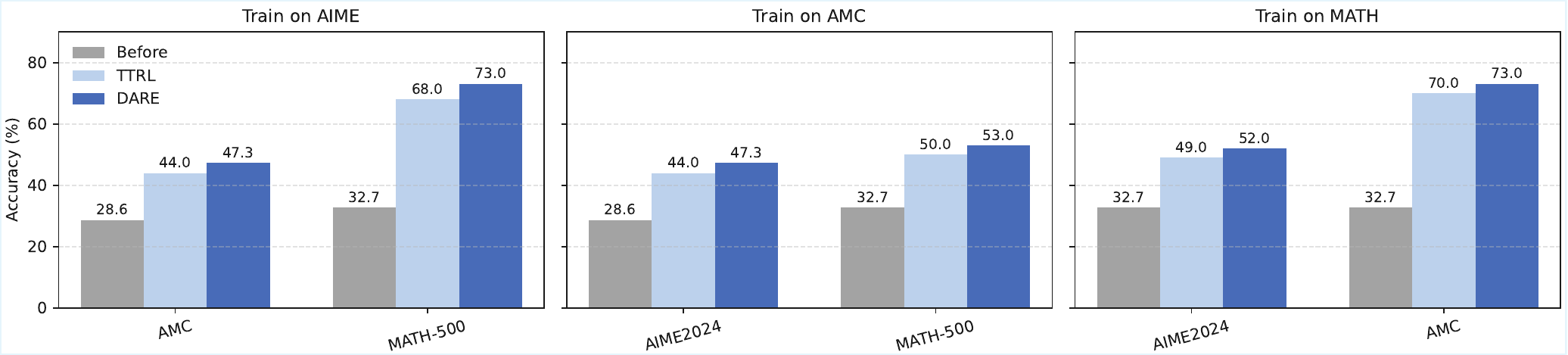}
\caption{\textbf{OOD generalization of Qwen2.5-Math-1.5B.} 
Each subfigure shows evaluation on OOD benchmarks after adaptation on a training set. 
Bars indicate  pass@1 accuracy for the original model, TTRL, and DARE, with DARE consistently improving performance.}

    \label{fig:ood_qwen25_ps_ttrl}
\end{figure*}

% -------------------------------
% 第二张图：Qwen3-1.7B
% -------------------------------
\begin{figure*}[t]
    \centering
    \includegraphics[width=\textwidth]{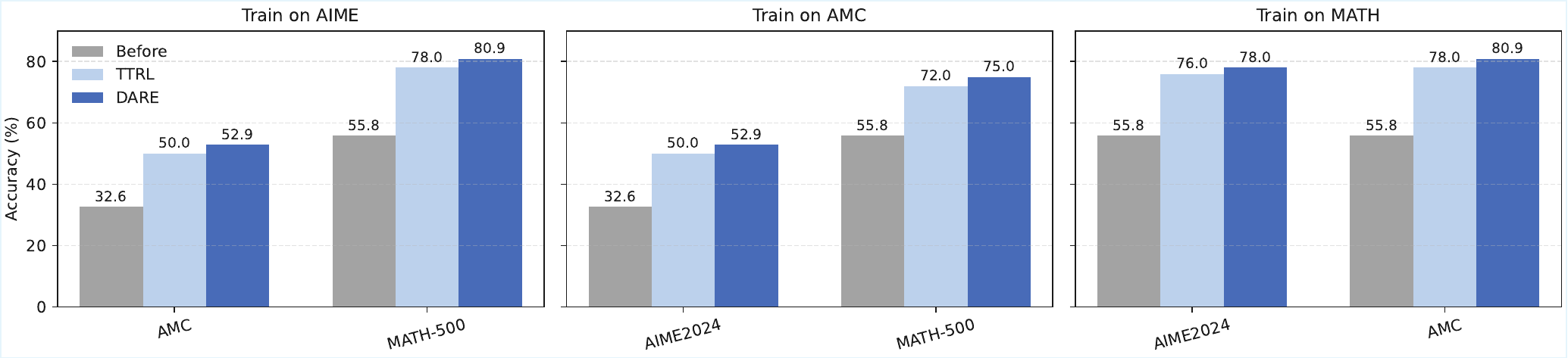}
    \caption{\textbf{OOD generalization of Qwen3-1.7B.} 
Each subfigure shows evaluation on OOD benchmarks after adaptation on a training set. 
Bars indicate pass@1 accuracy for the original model, TTRL, and DARE, with DARE consistently improving performance.}
    \label{fig:ood_qwen3_ps_ttrl}
\end{figure*}

\section{Experiments}

\paragraph{Evaluation Setup}
We evaluate (DARE on each benchmark independently. 
Unless specified, the maximum generation length is 3,072 tokens. 
We report pass@1 under stochastic decoding, sampling multiple actions per problem with temperature 1.0 and top-$p$ sampling ($p=0.95$).

\paragraph{Benchmarks and Models}
Experiments cover five benchmarks across three reasoning domains: \textit{general reasoning} (MMLU-Pro) \cite{wang2024mmluprorobustchallengingmultitask}, \textit{mathematical reasoning} (MATH-500, AIME~2024, AMC) \cite{li2024numinamath}, and \textit{scientific reasoning} (GPQA) \cite{rein2024gpqa}. 
We evaluate two backbone models: \texttt{Qwen2.5-Math-1.5B} and \texttt{Qwen3-1.7B} \cite{yang2024qwen2,yang2025qwen3}, covering both math-specialized and general-purpose architectures.

\paragraph{Baselines}
We compare DARE with three categories of baselines: 
(1) \textit{Prompting (training-free)}: raw backbone and Chain-of-Thought (CoT) \cite{COT}; 
(2) \textit{Reinforcement learning}: GRPO~\citep{shao2024deepseekmath}, REINFORCE~\citep{Williams1992SimpleSG}, and REINFORCE++~\citep{REINFORCE+}.  The three baselines are trained and evaluated on each dataset’s respective splits; due to AIME24’s limited size, MATH-trained models are evaluated on it directly.
(3) \textit{Test-time adaptation}: INTUITOR~\cite{intuitor}, RLPR~\cite{RLPR}, CO-REWARDING-I~\cite{coreward}, and TTRL~\cite{zuo2025ttrl}, which self-improve using self-generated signals.  
All methods share the same decoding budget and sampling strategy for fair comparison.

% \noindent\textbf{Implementation Details}
% We implement DARE on top of GRPO~\citep{shao2024deepseekmath} and apply it independently to each benchmark.
% All rollouts are generated with temperature $1.0$.
% We use AdamW with a cosine schedule and a peak learning rate of $5 \times 10^{-7}$.
% The shaping coefficient is fixed to $\alpha = 0.1$.
% For each iteration, we sample $64$ responses per prompt for reward estimation and use $32$ for policy updates.
% All experiments are conducted on 8 NVIDIA A100 GPUs (80GB).

\subsection{Main Results} Table~\ref{tab:exp} reports the main results on five benchmarks across two backbone models, comparing prompt-based methods, offline reinforcement learning, existing test-time scaling approaches, and DARE. 

\noindent\textbf{Overall Performance} Across both backbones and all benchmarks, DARE achieves the best average performance. On \texttt{Qwen2.5-Math-1.5B}, DARE raises the average from 41.56 (TTRL) to 44.20, with gains on all five benchmarks. On \texttt{Qwen3-1.7B}, it increases the average from 48.64 to 50.62, establishing new state-of-the-art among test-time scaling methods. These results show that probability-shaped rewards provide a more effective learning signal than majority-vote rewards across reasoning domains. 

\noindent\textbf{Gains across Reasoning Domains} DARE improves all three task categories. For \textit{general reasoning} (MMLU-Pro), it outperforms TTRL by +3.3 and +1.9 points on Qwen2.5-Math-1.5B and Qwen3-1.7B, respectively. For \textit{mathematical reasoning}, gains are most pronounced on AIME~2024, with +4.0 and +2.3 improvements, reflecting its effectiveness in correcting uncertain predictions. On \textit{scientific reasoning} (GPQA), DARE consistently surpasses all baselines, showing that distribution-aware rewards generalize beyond mathematical tasks.

\noindent\textbf{Comparison with Reinforcement Learning and Test-Time Scaling} Compared with offline RL methods, DARE achieves higher performance without extra training data; on Qwen3-1.7B, it beats REINFORCE++ by +2.4 points. Among test-time scaling methods, DARE surpasses INTUITOR, RLPR, CO-REWARDING-I, and standard TTRL on all benchmarks. While TTRL already improves over RL baselines, DARE adds 1.6--2.6 points in average performance, highlighting the benefit of exploiting rollout-level distributional structure. Largest gains appear on challenging benchmarks like AIME~2024, where majority-vote rewards are prone to entropy collapse, confirming that probability-shaped reward estimation is a principled and effective alternative to TTRL.

\subsection{Out-of-Distribution (OOD) Generalization}

To assess whether test-time adaptation generalizes beyond the benchmark used for adaptation, we conduct an out-of-distribution (OOD) evaluation across multiple reasoning datasets.
Figures~\ref{fig:ood_qwen25_ps_ttrl} and~\ref{fig:ood_qwen3_ps_ttrl} report results for Qwen2.5-Math-1.5B and Qwen3-1.7B, respectively, comparing the original backbone model (Before), standard TTRL, and DARE.

The results reveal two consistent trends.
First, standard TTRL improves performance on almost all unseen benchmarks relative to the original model, confirming that test-time adaptation learns transferable behaviors rather than overfitting to the adaptation benchmark. Second, DARE further improves upon standard TTRL across nearly all OOD settings, typically by 2--5 points, demonstrating that probability-shaped rewards provide a more informative and stable learning signal than majority-vote rewards.
By preserving trace-level uncertainty and leveraging the full rollout distribution, DARE enables more reliable policy updates and stronger generalization.

For example, when adapting on AIME 2024 and evaluating on MATH-500, DARE consistently outperforms standard TTRL by several points.
Similar gains are observed when adapting on AMC or MATH-500, indicating that the advantage of DARE is robust across adaptation scenarios.

Overall, these results show that DARE substantially enhances the OOD generalization of test-time reinforcement learning, leading to more reliable cross-benchmark performance than standard TTRL.

\begin{table}[t]
\centering
\small
\renewcommand{\arraystretch}{1.2}
\setlength{\tabcolsep}{8pt}

\begin{tabular}{lcc}
\toprule
\textbf{Model / Variant} & \textbf{AIME 2024} & \textbf{AMC} \\
\midrule

\multicolumn{3}{c}{\textsc{Qwen2.5-Math-1.5B}} \\
\midrule
Raw Model 
& 7.7 & 28.6 \\

+ Distribution Reward 
& 16.6 \,(\,+8.9) & 48.0 \,(\,+19.4) \\

+ Distribution + Bonus 
& 17.2 \,(\,+9.5) & 48.1 \,(\,+19.5) \\

+ Distribution + Prune 
& 17.5 \,(\,+9.8) & 49.3 \,(\,+20.7) \\

\rowcolor{blue!12}
\textbf{+ DARE} 
& \textbf{19.8 \,(\,+12.1)} & \textbf{50.2 \,(\,+21.6)} \\

\midrule
\multicolumn{3}{c}{\textsc{Qwen3-1.7B}} \\
\midrule
Raw Model 
& 11.5 & 32.6 \\

+ Distribution Reward 
& 24.7 \,(\,+13.2) & 53.8 \,(\,+21.2) \\

+ Distribution + Bonus 
& 24.3 \,(\,+12.8) & 54.9 \,(\,+22.3) \\

+ Distribution + Prune 
& 25.6 \,(\,+14.1) & 54.8 \,(\,+22.2) \\

\rowcolor{blue!12}
\textbf{+ DARE} 
& \textbf{26.3 \,(\,+14.8)} & \textbf{55.7 \,(\,+23.1)} \\

\bottomrule
\end{tabular}

\caption{Ablation results of DARE on AIME~2024 and AMC. Each row reports the performance after adding one component to the raw model. Numbers in parentheses denote absolute improvements over the corresponding raw baseline within each block.}

\label{tab:ablation}
\end{table}

\subsection{Ablation Study}

We conduct an ablation study to analyze the contribution of each component in DARE by progressively adding distribution-based reward, exploration bonus, and distribution pruning on top of the raw model (Table~\ref{tab:ablation}).  
Across both backbones and benchmarks, all components yield consistent improvements, and the full model achieves the best performance, demonstrating the effectiveness and complementarity of the proposed design.

We first observe that distribution-based reward provides the dominant gain.  
For example, on Qwen2.5-Math-1.5B, it improves AIME 2024 from 7.7 to 16.6 and AMC from 28.6 to 48.0, and similar trends are observed on Qwen3-1.7B.  
This indicates that reward estimation is the primary performance bottleneck in test-time reinforcement learning, and distribution-aware estimation constitutes the core contributor to performance improvements.

Building on this foundation, both exploration bonus and pruning provide complementary benefits by encouraging rare but valid rollouts and suppressing noisy samples.  
Combining all components yields the largest gains across all settings, confirming that DARE effectively integrates probability shaping, exploration control, and noise reduction into a unified reward estimation framework.

\begin{figure}[htbp]
    \centering
    \includegraphics[width=0.48\textwidth]{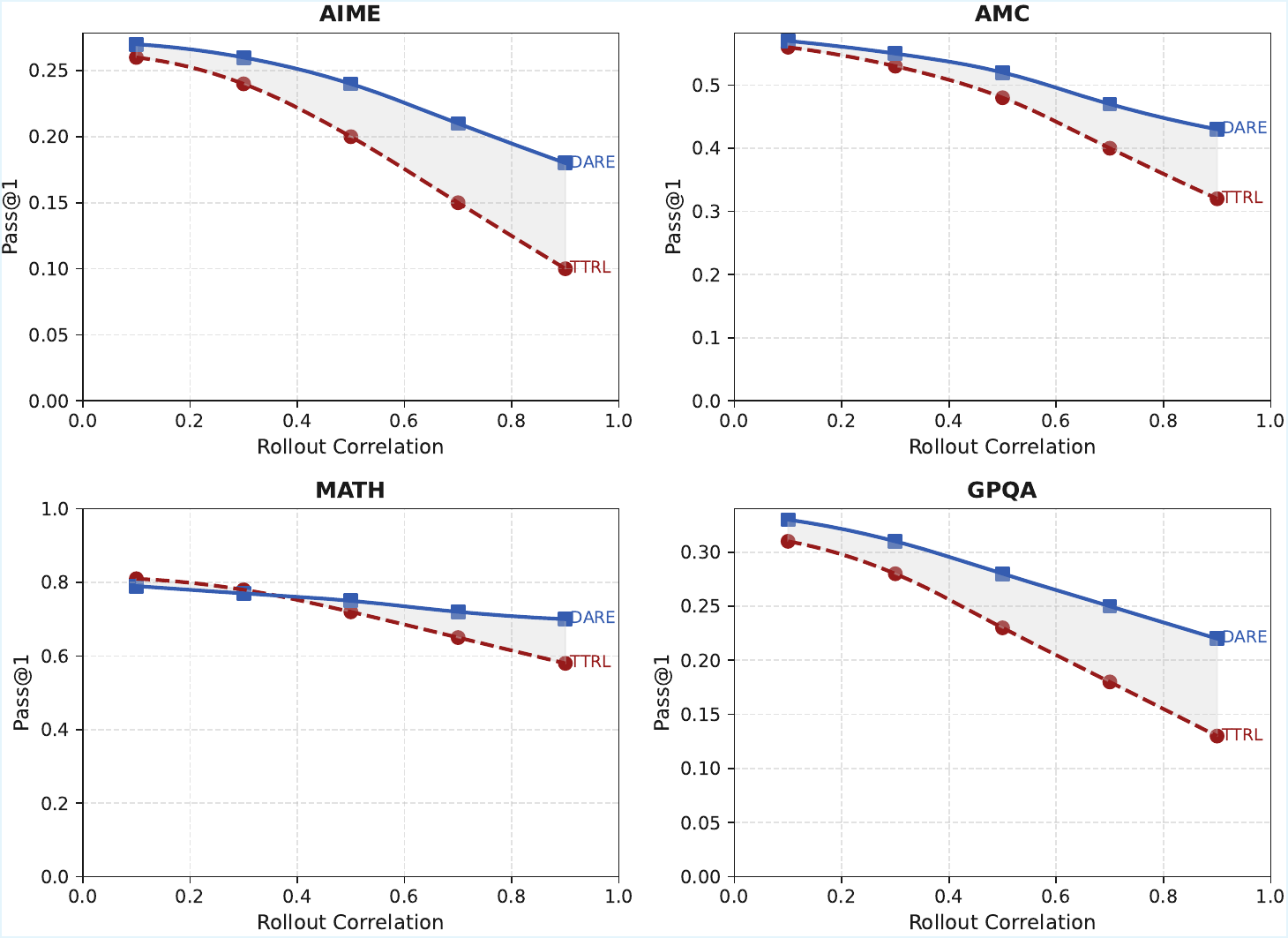}
   \caption{\textbf{Impact of rollout correlation on Qwen3-1.7B.} 
The x-axis represents \textit{Rollout Correlation}, and the y-axis shows \textit{Pass@1}. 
TTRL performance drops sharply with increasing correlation, while DARE degrades smoothly, demonstrating robustness.}

    \label{fig:correlation_performance}
\end{figure}

\subsection{Rollout Correlation Analysis}

To study the impact of correlated rollouts on reward estimation,
we adjust sampling temperature and decoding hyperparameters, which affect output diversity.
Rather than estimating latent correlation directly, we use an operational proxy: \emph{rollout correlation}, defined as the average pairwise token-level overlap between sampled rollouts for the same input, averaged across all pairs. 
This metric does not recover true statistical correlation but provides a reproducible measure of sample redundancy: higher similarity indicates more correlated rollouts.

Figure~\ref{fig:correlation_performance} plots model performance against this empirical similarity.
TTRL degrade rapidly as similarity increases, while DARE declines more gradually.
This aligns with our theoretical analysis: MV treats majority rollout correct, thereby introducing systematic bias; this bias is exacerbated when rollouts are more correlated, leading to more severe confirmation collapse. In contrast, DARE operates on the empirical rollout distribution and provide an more reliable reward estimation, effectively mitigating this issue.

\begin{figure*}[t]
    \centering
    \begin{subfigure}[t]{0.46\textwidth}
        \centering
        \includegraphics[width=\linewidth]{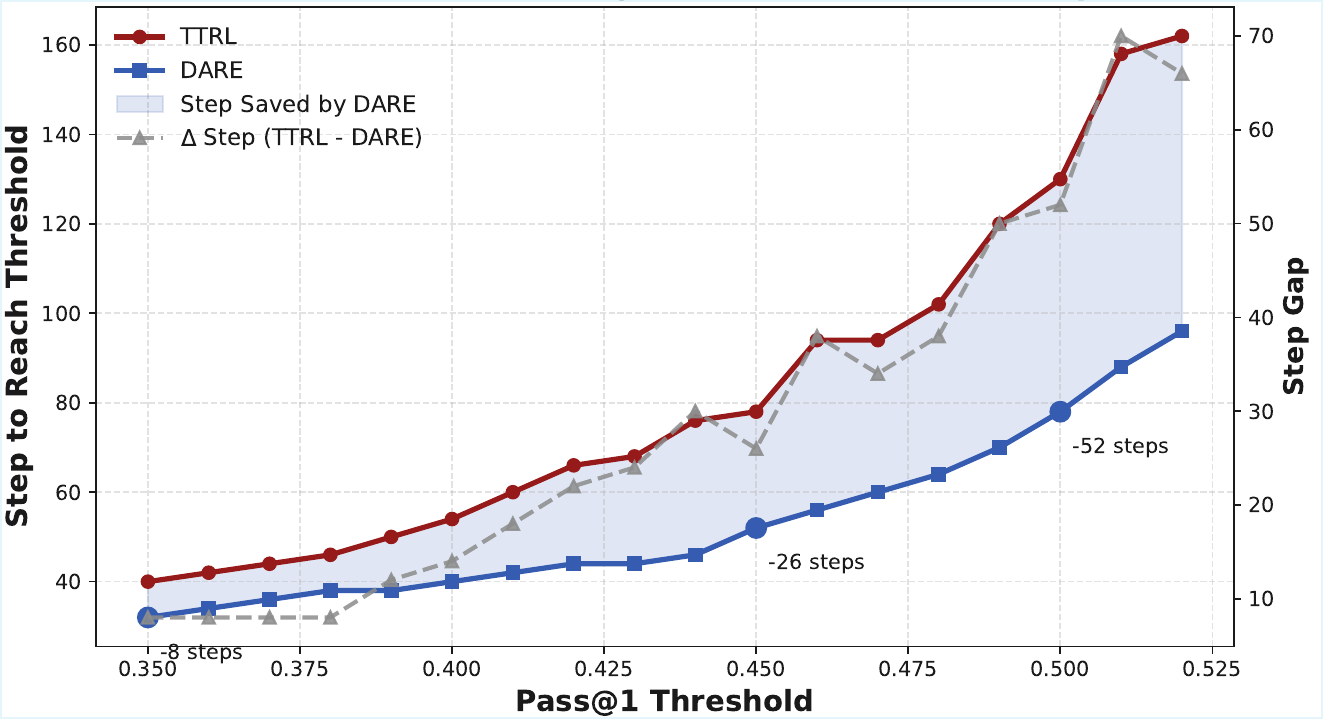}
        \caption{AMC (Qwen3-1.7B)}
        \label{fig:conv_amc_17b}
    \end{subfigure}
    \hfill
    \begin{subfigure}[t]{0.46\textwidth}
        \centering
        \includegraphics[width=\linewidth]{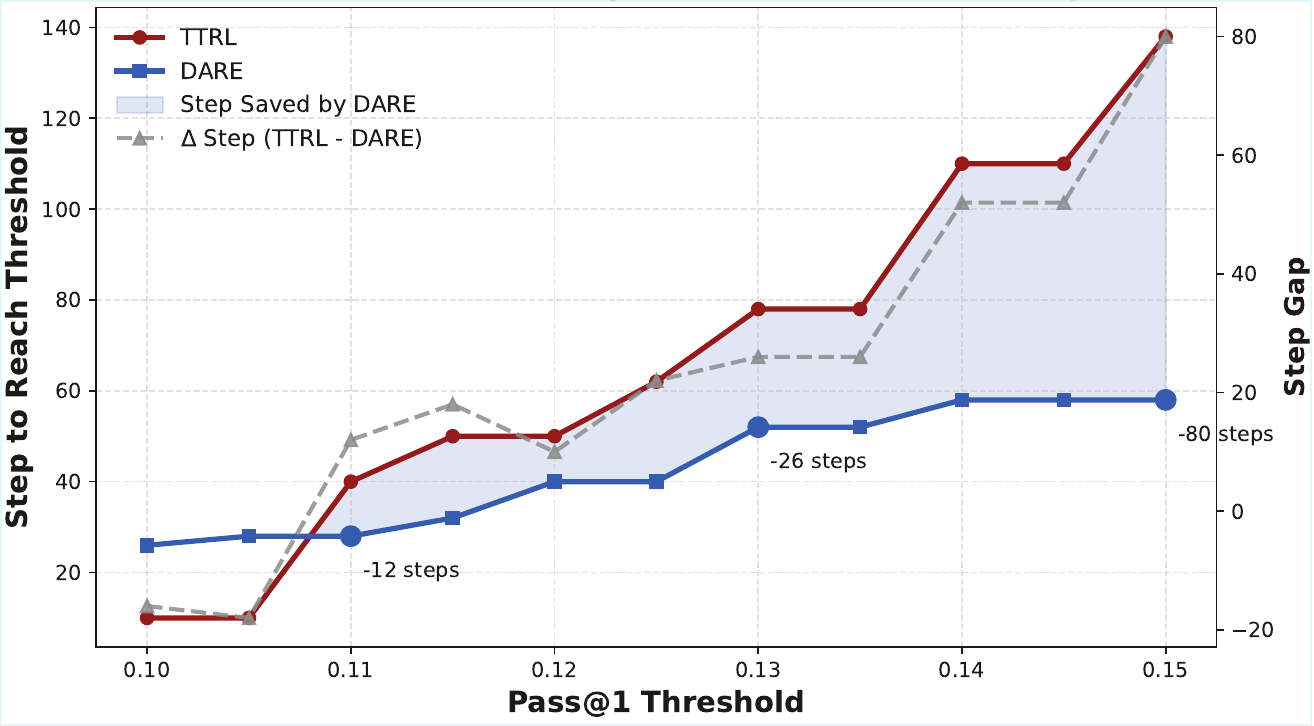}
        \caption{AIME (Qwen2.5-Math-1.5B)}
        \label{fig:conv_aime_15b}
    \end{subfigure}

    \caption{\textbf{Convergence speed comparison of TTRL and DARE.}
    The x-axis denotes pass@1 thresholds, and the y-axis indicates the minimum number of steps required to reach each threshold.
    Across both AMC and AIME benchmarks, DARE consistently achieves the same performance with fewer steps, demonstrating faster test-time adaptation.}
    \label{fig:convergence_speed}
\end{figure*}

\subsection{Training Dynamics and Convergence Analysis}

To evaluate the test-time adaptation efficiency of DARE, we analyze its convergence speed in terms of the number of update steps required to reach predefined accuracy thresholds.
Figure~\ref{fig:convergence_speed} reports, for each accuracy level, the minimum number of steps needed to reach that threshold, rather than accuracy as a function of steps, which directly reflects adaptation efficiency.

Across both AMC and AIME benchmarks, DARE consistently requires fewer steps than standard TTRL to achieve the same accuracy, indicating faster convergence and more sample-efficient adaptation.
This advantage demonstrates that DARE can achieve moderate-to-high accuracy levels substantially earlier than TTRL, highlighting its robustness in diverse settings.

These results suggest that DARE improves the efficiency of each update by preserving trace-level uncertainty and leveraging the full empirical rollout distribution, which provides richer and less biased reward signals than MV-based reward.
As a result, DARE achieves higher information gain per step, stabilizes early-stage optimization, and accelerates policy learning, leading to faster test-time adaptation.

\section{Related Work}
\paragraph{Enhancing Reasoning in Large Language Models.}
A major line of recent research improves the reasoning ability of large language models through reinforcement learning with verifiable outcomes (RLVR) \cite{jaech2024openai, guo2025deepseek, yang2025qwen3, dai2025r1}. 
In this paradigm, learning is guided by tasks whose actions can be automatically validated, most notably in mathematical reasoning and program synthesis \cite{zeng2025simplerl, wang2025adareasoner, cui2025entropy, dai2025cde}. 
The availability of reliable verifiers enables stable reward signals and has led to substantial gains in reasoning performance. 
However, this assumption fundamentally limits the scope of RLVR, as many real-world reasoning problems lack deterministic or easily checkable answers \cite{zhao2025one, zhou2025reinforcing, zhou2024defending,zhi2025reinventing}. 
Addressing this limitation, our work explores reasoning enhancement beyond verifier-dependent settings, targeting more general scenarios where explicit correctness signals are unavailable.

\paragraph{Label-Free Adaptation and Self-Improvement.}
Recent work studies label-free adaptation, enabling models to self-improve under distribution shift by constructing proxy rewards from their own generations.
Existing approaches broadly fall into two lines.
The first exploits intrinsic confidence signals, reinforcing predictions that are internally consistent or low-uncertainty, typically via entropy- or agreement-based criteria. \citep{prabhudesai2025maximizing, agarwal2025unreasonable, zhao2025learning, zhang2025right}.
The second line, which our work most closely relates to, bootstraps supervision from population agreement.
Test-Time Reinforcement Learning (TTRL) selects a majority-voted answer from multiple rollouts as a pseudo-label \citep{zuo2025ttrl}, and subsequent works refine this paradigm along different dimensions.
Evol-RL \citep{zhou2025evolvinglanguagemodelslabels} iteratively updates the policy using MV-based novelty labels, RESTRAIN \cite{yu2025restrainspuriousvotessignals} exploiting signals from the model's own answer distribution to penalize overconfidence and reward consistency to mitigate MV's collapse, while more recent methods enrich the majority signal with paraphrased views, step-wise confidence, subgroup voting, and calibrated decisiveness rewards
\citep{selfharmony, wang2025scope, xing2025compass,wu2025spinetokenselectivetesttimereinforcement}.
Despite their differences, these methods share a common assumption: reward quality is inferred from consensus, with majority agreement as the learning target.
In contrast, we identify this \emph{MV drawbacks} as a fundamental limitation of consensus-based TTRL and redesign the learning target from a distributional view.

\section{Conclusion}
In this work, we theoretically analyze reward estimation in test-time reinforcement learning and show that majority-vote rewards provide a fragile learning signal by reducing rollouts into single label and inducing systematic bias.
To address this, we propose \textbf{DARE}, which shifts reward estimation from point-level consensus to the empirical distribution.
This distribution-aware view yields more informative and robust reward signals, improving stability and final performance across multiple reasoning benchmarks.
More broadly, our results suggest that the distribution space provides a principled foundation for reward shaping in test-time adaptation, opening avenues for richer distributional designs and uncertainty-aware reward estimation in the future.

\section{Impact Statement}

This paper presents work whose goal is to advance the field of machine learning. 
There are many potential societal consequences of our work, none of which we feel must be specifically highlighted here.

\bibliography{example_paper}

@article{zhi2025reinventing,
  title={Reinventing Clinical Dialogue: Agentic Paradigms for LLM Enabled Healthcare Communication},
  author={Zhi, Xiaoquan and Zhao, Hongke and Wu, Likang and Zhao, Chuang and Zhu, Hengshu},
  journal={arXiv preprint arXiv:2512.01453},
  year={2025}
}

@misc{yu2025restrainspuriousvotessignals,
      title={RESTRAIN: From Spurious Votes to Signals -- Self-Driven RL with Self-Penalization}, 
      author={Zhaoning Yu and Will Su and Leitian Tao and Haozhu Wang and Aashu Singh and Hanchao Yu and Jianyu Wang and Hongyang Gao and Weizhe Yuan and Jason Weston and Ping Yu and Jing Xu},
      year={2025},
      eprint={2510.02172},
      archivePrefix={arXiv},
      primaryClass={cs.CL},
      url={https://arxiv.org/abs/2510.02172}, 
}

@article{Williams1992SimpleSG,
  title={Simple Statistical Gradient-Following Algorithms for Connectionist Reinforcement Learning},
  author={Ronald J. Williams},
  journal={Machine Learning},
  year={1992},
  volume={8},
  pages={229-256}
}

@misc{REINFORCE+,
      title={REINFORCE++: Stabilizing Critic-Free Policy Optimization with Global Advantage Normalization}, 
      author={Jian Hu and Jason Klein Liu and Haotian Xu and Wei Shen},
      year={2025},
      eprint={2501.03262},
      archivePrefix={arXiv},
      primaryClass={cs.CL},
      url={https://arxiv.org/abs/2501.03262}, 
}

@misc{COT,
      title={Chain-of-Thought Prompting Elicits Reasoning in Large Language Models}, 
      author={Jason Wei and Xuezhi Wang and Dale Schuurmans and Maarten Bosma and Brian Ichter and Fei Xia and Ed Chi and Quoc Le and Denny Zhou},
      year={2023},
      eprint={2201.11903},
      archivePrefix={arXiv},
      primaryClass={cs.CL},
      url={https://arxiv.org/abs/2201.11903}, 
}

@misc{wang2024mmluprorobustchallengingmultitask,
      title={MMLU-Pro: A More Robust and Challenging Multi-Task Language Understanding Benchmark}, 
      author={Yubo Wang and Xueguang Ma and Ge Zhang and Yuansheng Ni and Abhranil Chandra and Shiguang Guo and Weiming Ren and Aaran Arulraj and Xuan He and Ziyan Jiang and Tianle Li and Max Ku and Kai Wang and Alex Zhuang and Rongqi Fan and Xiang Yue and Wenhu Chen},
      year={2024},
      eprint={2406.01574},
      archivePrefix={arXiv},
      primaryClass={cs.CL},
      url={https://arxiv.org/abs/2406.01574}, 
}

@misc{chang2025steplevelverifierguidedhybridtesttime,
      title={Step-level Verifier-guided Hybrid Test-Time Scaling for Large Language Models}, 
      author={Kaiyan Chang and Yonghao Shi and Chenglong Wang and Hang Zhou and Chi Hu and Xiaoqian Liu and Yingfeng Luo and Yuan Ge and Tong Xiao and Jingbo Zhu},
      year={2025},
      eprint={2507.15512},
      archivePrefix={arXiv},
      primaryClass={cs.CL},
      url={https://arxiv.org/abs/2507.15512}, 
}

@misc{coreward,
      title={Co-rewarding: Stable Self-supervised RL for Eliciting Reasoning in Large Language Models}, 
      author={Zizhuo Zhang and Jianing Zhu and Xinmu Ge and Zihua Zhao and Zhanke Zhou and Xuan Li and Xiao Feng and Jiangchao Yao and Bo Han},
      year={2025},
      eprint={2508.00410},
      archivePrefix={arXiv},
      primaryClass={cs.LG},
      url={https://arxiv.org/abs/2508.00410}, 
}

@misc{RLPR,
      title={RLPR: Extrapolating RLVR to General Domains without Verifiers}, 
      author={Tianyu Yu and Bo Ji and Shouli Wang and Shu Yao and Zefan Wang and Ganqu Cui and Lifan Yuan and Ning Ding and Yuan Yao and Zhiyuan Liu and Maosong Sun and Tat-Seng Chua},
      year={2025},
      eprint={2506.18254},
      archivePrefix={arXiv},
      primaryClass={cs.LG},
      url={https://arxiv.org/abs/2506.18254}, 
}

@misc{intuitor,
      title={Learning to Reason without External Rewards}, 
      author={Xuandong Zhao and Zhewei Kang and Aosong Feng and Sergey Levine and Dawn Song},
      year={2025},
      eprint={2505.19590},
      archivePrefix={arXiv},
      primaryClass={cs.LG},
      url={https://arxiv.org/abs/2505.19590}, 
}

@misc{selfharmony,
      title={Self-Harmony: Learning to Harmonize Self-Supervision and Self-Play in Test-Time Reinforcement Learning}, 
      author={Ru Wang and Wei Huang and Qi Cao and Yusuke Iwasawa and Yutaka Matsuo and Jiaxian Guo},
      year={2025},
      eprint={2511.01191},
      archivePrefix={arXiv},
      primaryClass={cs.CL},
      url={https://arxiv.org/abs/2511.01191}, 
}

@misc{xing2025compass,
      title={Rewarding the Journey, Not Just the Destination: A Composite Path and Answer Self-Scoring Reward Mechanism for Test-Time Reinforcement Learning}, 
      author={Jingyu Xing and Chenwei Tang and Xinyu Liu and Deng Xiong and Shudong Huang and Wei Ju and Jiancheng Lv and Ziyue Qiao},
      year={2025},
      eprint={2510.17923},
      archivePrefix={arXiv},
      primaryClass={cs.LG},
      url={https://arxiv.org/abs/2510.17923}, 
}

@misc{wang2025scope,
      title={Beyond Majority Voting: Towards Fine-grained and More Reliable Reward Signal for Test-Time Reinforcement Learning}, 
      author={Weiqin Wang and Yile Wang and Kehao Chen and Hui Huang},
      year={2025},
      eprint={2512.15146},
      archivePrefix={arXiv},
      primaryClass={cs.CL},
      url={https://arxiv.org/abs/2512.15146}, 
}

@article{zuo2025ttrl,
  title={Ttrl: Test-time reinforcement learning},
  author={Zuo, Yuxin and Zhang, Kaiyan and Sheng, Li and Qu, Shang and Cui, Ganqu and Zhu, Xuekai and Li, Haozhan and Zhang, Yuchen and Long, Xinwei and Hua, Ermo and others},
  journal={arXiv preprint arXiv:2504.16084},
  year={2025}
}

@article{jaech2024openai,
  title={Openai o1 system card},
  author={Jaech, Aaron and Kalai, Adam and Lerer, Adam and Richardson, Adam and El-Kishky, Ahmed and Low, Aiden and Helyar, Alec and Madry, Aleksander and Beutel, Alex and Carney, Alex and others},
  journal={arXiv preprint arXiv:2412.16720},
  year={2024}
}

@article{guo2025deepseek,
  title={Deepseek-r1: Incentivizing reasoning capability in llms via reinforcement learning},
  author={Guo, Daya and Yang, Dejian and Zhang, Haowei and Song, Junxiao and Zhang, Ruoyu and Xu, Runxin and Zhu, Qihao and Ma, Shirong and Wang, Peiyi and Bi, Xiao and others},
  journal={arXiv preprint arXiv:2501.12948},
  year={2025}
}

@article{dai2025r1,
  title={R1-RE: Cross-Domain Relation Extraction with RLVR},
  author={Dai, Runpeng and Zheng, Tong and Yang, Run and Yu, Kaixian and Zhu, Hongtu},
  journal={arXiv preprint arXiv:2507.04642},
  year={2025}
}

@article{zeng2025simplerl,
  title={Simplerl-zoo: Investigating and taming zero reinforcement learning for open base models in the wild},
  author={Zeng, Weihao and Huang, Yuzhen and Liu, Qian and Liu, Wei and He, Keqing and Ma, Zejun and He, Junxian},
  journal={arXiv preprint arXiv:2503.18892},
  year={2025}
}

@article{wang2025adareasoner,
  title={AdaReasoner: Adaptive Reasoning Enables More Flexible Thinking},
  author={Wang, Xiangqi and Huang, Yue and Wang, Yanbo and Luo, Xiaonan and Guo, Kehan and Zhou, Yujun and Zhang, Xiangliang},
  journal={arXiv preprint arXiv:2505.17312},
  year={2025}
}

@article{cui2025entropy,
  title={The entropy mechanism of reinforcement learning for reasoning language models},
  author={Cui, Ganqu and Zhang, Yuchen and Chen, Jiacheng and Yuan, Lifan and Wang, Zhi and Zuo, Yuxin and Li, Haozhan and Fan, Yuchen and Chen, Huayu and Chen, Weize and others},
  journal={arXiv preprint arXiv:2505.22617},
  year={2025}
}

@article{dai2025cde,
  title={CDE: Curiosity-Driven Exploration for Efficient Reinforcement Learning in Large Language Models},
  author={Dai, Runpeng and Song, Linfeng and Liu, Haolin and Liang, Zhenwen and Yu, Dian and Mi, Haitao and Tu, Zhaopeng and Liu, Rui and Zheng, Tong and Zhu, Hongtu and others},
  journal={arXiv preprint arXiv:2509.09675},
  year={2025}
}

@article{zhao2025one,
  title={One Token to Fool LLM-as-a-Judge},
  author={Zhao, Yulai and Liu, Haolin and Yu, Dian and Kung, SY and Mi, Haitao and Yu, Dong},
  journal={arXiv preprint arXiv:2507.08794},
  year={2025}
}

@article{zhou2025reinforcing,
  title={Reinforcing General Reasoning without Verifiers},
  author={Zhou, Xiangxin and Liu, Zichen and Sims, Anya and Wang, Haonan and Pang, Tianyu and Li, Chongxuan and Wang, Liang and Lin, Min and Du, Chao},
  journal={arXiv preprint arXiv:2505.21493},
  year={2025}
}

@article{zhou2024defending,
  title={Defending jailbreak prompts via in-context adversarial game},
  author={Zhou, Yujun and Han, Yufei and Zhuang, Haomin and Guo, Kehan and Liang, Zhenwen and Bao, Hongyan and Zhang, Xiangliang},
  journal={arXiv preprint arXiv:2402.13148},
  year={2024}
}

@article{prabhudesai2025maximizing,
  title={Maximizing Confidence Alone Improves Reasoning},
  author={Prabhudesai, Mihir and Chen, Lili and Ippoliti, Alex and Fragkiadaki, Katerina and Liu, Hao and Pathak, Deepak},
  journal={arXiv preprint arXiv:2505.22660},
  year={2025}
}

@article{agarwal2025unreasonable,
  title={The unreasonable effectiveness of entropy minimization in llm reasoning},
  author={Agarwal, Shivam and Zhang, Zimin and Yuan, Lifan and Han, Jiawei and Peng, Hao},
  journal={arXiv preprint arXiv:2505.15134},
  year={2025}
}

@article{zhao2025learning,
  title={Learning to reason without external rewards},
  author={Zhao, Xuandong and Kang, Zhewei and Feng, Aosong and Levine, Sergey and Song, Dawn},
  journal={arXiv preprint arXiv:2505.19590},
  year={2025}
}

@article{zhang2025right,
  title={Right question is already half the answer: Fully unsupervised llm reasoning incentivization},
  author={Zhang, Qingyang and Wu, Haitao and Zhang, Changqing and Zhao, Peilin and Bian, Yatao},
  journal={arXiv preprint arXiv:2504.05812},
  year={2025}
}

@misc{zhou2025evolvinglanguagemodelslabels,
      title={Evolving Language Models without Labels: Majority Drives Selection, Novelty Promotes Variation}, 
      author={Yujun Zhou and Zhenwen Liang and Haolin Liu and Wenhao Yu and Kishan Panaganti and Linfeng Song and Dian Yu and Xiangliang Zhang and Haitao Mi and Dong Yu},
      year={2025},
      eprint={2509.15194},
      archivePrefix={arXiv},
      primaryClass={cs.LG},
      url={https://arxiv.org/abs/2509.15194}, 
}

@misc{wu2025spinetokenselectivetesttimereinforcement,
      title={SPINE: Token-Selective Test-Time Reinforcement Learning with Entropy-Band Regularization}, 
      author={Jianghao Wu and Yasmeen George and Jin Ye and Yicheng Wu and Daniel F. Schmidt and Jianfei Cai},
      year={2025},
      eprint={2511.17938},
      archivePrefix={arXiv},
      primaryClass={cs.CL},
      url={https://arxiv.org/abs/2511.17938}, 
}

@article{ahn2024large,
  title={Large language models for mathematical reasoning: Progresses and challenges},
  author={Ahn, Janice and Verma, Rishu and Lou, Renze and Liu, Di and Zhang, Rui and Yin, Wenpeng},
  journal={arXiv preprint arXiv:2402.00157},
  year={2024}
}

@article{plaat2024reasoning,
  title={Reasoning with large language models, a survey},
  author={Plaat, Aske and Wong, Annie and Verberne, Suzan and Broekens, Joost and van Stein, Niki and Back, Thomas},
  journal={arXiv preprint arXiv:2407.11511},
  year={2024}
}

@article{li2025fundamental,
  title={Fundamental capabilities and applications of large language models: A survey},
  author={Li, Jiawei and Gao, Yang and Yang, Yizhe and Bai, Yu and Zhou, Xiaofeng and Li, Yinghao and Sun, Huashan and Liu, Yuhang and Si, Xingpeng and Ye, Yuhao and others},
  journal={ACM Computing Surveys},
  year={2025},
  publisher={ACM New York, NY}
}

@inproceedings{huang2023towards,
  title={Towards reasoning in large language models: A survey},
  author={Huang, Jie and Chang, Kevin Chen-Chuan},
  booktitle={Findings of the association for computational linguistics: ACL 2023},
  pages={1049--1065},
  year={2023}
}

@article{chen2024more,
  title={Are more llm calls all you need? towards the scaling properties of compound ai systems},
  author={Chen, Lingjiao and Davis, Jared and Hanin, Boris and Bailis, Peter and Stoica, Ion and Zaharia, Matei and Zou, James},
  journal={Advances in Neural Information Processing Systems},
  volume={37},
  pages={45767--45790},
  year={2024}
}

@article{majumdar2024generative,
  title={Generative AI voting: fair collective choice is resilient to LLM biases and inconsistencies},
  author={Majumdar, Srijoni and Elkind, Edith and Pournaras, Evangelos},
  journal={arXiv preprint arXiv:2406.11871},
  year={2024}
}

@inproceedings{gudatiny,
  title={TINY: Rethinking Selection Bias in LLMs: Quantification and Mitigation using Efficient Majority Voting},
  author={Guda, Blessed and Francis, Lawrence and Ashungafac, Gabrial Zencha and Joe-Wong, Carlee and Busogi, Moise},
  booktitle={ICLR Workshop: Quantify Uncertainty and Hallucination in Foundation Models: The Next Frontier in Reliable AI}
}

@article{yu2025pass,
  title={Pass@ k Metric for RLVR: A Diagnostic Tool of Exploration, But Not an Objective},
  author={Yu, Yang},
  journal={arXiv preprint arXiv:2511.16231},
  year={2025}
}

@article{rein2024gpqa,
  title   = {GPQA: A Graduate-Level Google-Proof Question Answering Benchmark},
  author  = {Rein, David and Hou, Betty Li and Stickland, Asa Cooper and Petty, Jackson and Pang, Richard Yuanzhe and Dirani, Julien and Michael, Julian and Bowman, Samuel R.},
  journal = {First Conference on Language Modeling},
  year    = {2024}
}

@article{li2024numinamath,
  title   = {NuminaMath: The Largest Public Dataset in AI4Maths with 860K Competition Math Problems and Solutions},
  author  = {Li, Jia and Beeching, Edward and Tunstall, Lewis and Lipkin, Ben and Soletskyi, Roman and Huang, Shengyi and Rasul, Kashif and Yu, Longhui and Jiang, Albert Q. and Shen, Ziju and others},
  journal = {Hugging Face Dataset},
  year    = {2024}
}

@article{yang2024qwen2,
  title={Qwen2. 5-math technical report: Toward mathematical expert model via self-improvement},
  author={Yang, An and Zhang, Beichen and Hui, Binyuan and Gao, Bofei and Yu, Bowen and Li, Chengpeng and Liu, Dayiheng and Tu, Jianhong and Zhou, Jingren and Lin, Junyang and others},
  journal={arXiv preprint arXiv:2409.12122},
  year={2024}
}

@article{yang2025qwen3,
  title={Qwen3 technical report},
  author={Yang, An and Li, Anfeng and Yang, Baosong and Zhang, Beichen and Hui, Binyuan and Zheng, Bo and Yu, Bowen and Gao, Chang and Huang, Chengen and Lv, Chenxu and others},
  journal={arXiv preprint arXiv:2505.09388},
  year={2025}
}

@article{shao2024deepseekmath,
  title={Deepseekmath: Pushing the limits of mathematical reasoning in open language models},
  author={Shao, Zhihong and Wang, Peiyi and Zhu, Qihao and Xu, Runxin and Song, Junxiao and Bi, Xiao and Zhang, Haowei and Zhang, Mingchuan and Li, YK and Wu, Yang and others},
  journal={arXiv preprint arXiv:2402.03300},
  year={2024}
}
\bibliographystyle{icml2026}

%%%%%%%%%%%%%%%%%%%%%%%%%%%%%%%%%%%%%%%%%%%%%%%%%%%%%%%%%%%%%%%%%%%%%%%%%%%%%%%
%%%%%%%%%%%%%%%%%%%%%%%%%%%%%%%%%%%%%%%%%%%%%%%%%%%%%%%%%%%%%%%%%%%%%%%%%%%%%%%
% APPENDIX
%%%%%%%%%%%%%%%%%%%%%%%%%%%%%%%%%%%%%%%%%%%%%%%%%%%%%%%%%%%%%%%%%%%%%%%%%%%%%%%
%%%%%%%%%%%%%%%%%%%%%%%%%%%%%%%%%%%%%%%%%%%%%%%%%%%%%%%%%%%%%%%%%%%%%%%%%%%%%%%
\newpage
\appendix
\onecolumn

\appendix
\section{Proofs of Theoretical Results}

\subsection{Proof of Theorem 2.1}

Let $Y \sim p(y)$ denote a random rollout with reward $R(Y) \in \{0,1\}$.  
Majority Voting defines a deterministic mapping from the rollout population
$P = (Y_1,\dots,Y_M)$ to a pseudo-label $\hat{y}(P)$ and an induced reward signal
\[
R_{\mathrm{MV}}(Y;P) = R\!\left(Y,\hat{y}(P)\right).
\]

Since $R(Y)$ is a deterministic function of $Y$, and $R_{\mathrm{MV}}(Y;P)$ is a deterministic function of $(Y,P)$,
we consider the joint random variables $(R(Y), Y, R_{\mathrm{MV}}(Y;P))$.

By construction, $R_{\mathrm{MV}}(Y;P)$ is obtained by applying a deterministic transformation
to the rollout population, which includes $Y$ as one component.  
Therefore, $R_{\mathrm{MV}}(Y;P)$ is a (possibly many-to-one) measurable function of $Y$ together with auxiliary variables
that are independent of $R(Y)$ given $Y$.

Applying the data processing inequality to the transformation
\[
Y \;\mapsto\; R_{\mathrm{MV}}(Y;P),
\]
we obtain
\[
I(R(Y); R_{\mathrm{MV}}(Y;P)) \;\le\; I(R(Y); Y).
\]

Equality holds if and only if $R_{\mathrm{MV}}(Y;P)$ is a sufficient statistic of $Y$ with respect to $R(Y)$,
that is, when the mapping preserves all reward-relevant information contained in $Y$.

This occurs only in the degenerate case where all rollouts with nonzero probability  share the same reward value.  
Whenever multiple outputs with distinct rewards have nonzero probability, the mapping
$Y \mapsto R_{\mathrm{MV}}(Y;P)$ is many-to-one and strictly discards reward-relevant information.  
Consequently,
\[
I(R(Y); R_{\mathrm{MV}}(Y;P)) \;<\; I(R(Y); Y),
\]
which completes the proof.
\qed

\subsection{Proof of Theorem 2.2}

Let $Z$ be a latent variable inducing correlation among rollouts, and let 
$\hat{y}_1,\dots,\hat{y}_M \mid Z \overset{\text{i.i.d.}}{\sim} p(\hat{y}\mid Z)$.  
Denote by $R(\hat{y}) \in \{0,1\}$ the true reward of an individual rollout, and define the marginal expected reward
\[
\mu \;=\; \mathbb{E}_{\hat{y} \sim p(\hat{y})}[R(\hat{y})].
\]

Majority Voting selects the most frequent outcome among $\{\hat{y}_m\}_{m=1}^M$, and assigns reward 
$\hat{R}_{\mathrm{MV}} = R(\hat{y}^*)$, where $\hat{y}^*$ denotes the modal rollout.

Conditioned on $Z$, as $M \to \infty$, the empirical distribution of rollouts converges almost surely to 
$p(\hat{y} \mid Z)$, and $\hat{y}^*$ converges to the conditional mode
\[
\hat{y}^*(Z) \;=\; \arg\max_{\hat{y}} p(\hat{y}\mid Z).
\]
Therefore, the asymptotic reward assigned by MV satisfies
\[
\hat{R}_{\mathrm{MV}} \;\xrightarrow[M\to\infty]{}\; R(\hat{y}^*(Z)).
\]

Taking expectation over $Z$, we obtain
\[
\mathbb{E}[\hat{R}_{\mathrm{MV}}] 
\;=\; \mathbb{E}_{Z}\big[ R(\hat{y}^*(Z)) \big].
\]
In contrast, the marginal objective is
\[
\mathbb{E}_{\hat{y} \sim p}[R(\hat{y})]
\;=\; \mathbb{E}_{Z}\, \mathbb{E}_{\hat{y} \sim p(\hat{y}\mid Z)}[R(\hat{y})].
\]

Unless the conditional mode $\hat{y}^*(Z)$ coincides almost surely with the maximizer of the marginal expected reward,
the two quantities differ:
\[
\mathbb{E}_{Z}\big[ R(\hat{y}^*(Z)) \big]
\;\neq\;
\mathbb{E}_{Z}\, \mathbb{E}_{\hat{y} \sim p(\hat{y}\mid Z)}[R(\hat{y})].
\]

Under positive correlation, $p(\hat{y}\mid Z)$ concentrates probability on latent-specific modes, causing MV to overweight 
conditional modes favored by particular realizations of $Z$ rather than the marginally optimal action.  
Hence,
\[
\mathrm{Bias}_{\mathrm{MV}} 
= \mathbb{E}[\hat{R}_{\mathrm{MV}}] - \mathbb{E}_{\hat{y} \sim p}[R(\hat{y})] \;\neq\; 0,
\]
which establishes that MV yields biased reward estimation under correlated rollouts.

\qed

\subsection{Proof of Proposition 2.3}

Under the latent-variable model, rollouts are generated as follows.  
A latent variable $Z \sim p(Z)$ is first sampled, and then each rollout is drawn conditionally independently as
\[
\hat{y}_k \mid Z \sim p(y \mid Z), \quad k = 1,\dots,M.
\]
The empirical distribution is defined by
\[
\hat{p}(y) = \frac{1}{M} \sum_{k=1}^{M} \mathbf{1}[\hat{y}_k = y].
\]

\paragraph{Step 1: Unbiased estimation of the marginal rollout distribution.}

Taking expectation with respect to the joint distribution over $Z$ and rollouts, we have
\[
\mathbb{E}[\hat{p}(y)]
= \frac{1}{M} \sum_{k=1}^{M} \mathbb{E}\!\left[ \mathbf{1}[\hat{y}_k = y] \right].
\]
By the law of total expectation,
\[
\mathbb{E}\!\left[ \mathbf{1}[\hat{y}_k = y] \right]
= \mathbb{E}_{Z} \, \mathbb{E}_{\hat{y}_k \mid Z} \!\left[ \mathbf{1}[\hat{y}_k = y] \right]
= \mathbb{E}_{Z} [ p(y \mid Z) ].
\]
Therefore,
\[
\mathbb{E}[\hat{p}(y)]
= \mathbb{E}_{Z} [ p(y \mid Z) ]
= p(y),
\]
which shows that the empirical probability $\hat{p}(y)$ is an unbiased estimator of the marginal rollout distribution $p(y)$.

\paragraph{Step 2: Induced proxy objective.}

The distribution-based reward assigns
\[
R_{\mathrm{dist}}(y) = g(\hat{p}(y)),
\]
where $g(\cdot)$ is a monotonic shaping function.  
Taking expectation yields the proxy objective
\[
J_{\mathrm{dist}}(\theta)
=
\mathbb{E}_{y \sim p_\theta(y)} [ g(p_\theta(y)) ].
\]
This objective depends only on the marginal rollout distribution $p_\theta(y)$ and is invariant to the latent variable $Z$.

\paragraph{Step 3: Policy gradient alignment with the marginal distribution.}

Consider the induced policy gradient:
\[
\nabla_\theta J_{\mathrm{dist}}(\theta)
=
\nabla_\theta \sum_y p_\theta(y) \, g(p_\theta(y)).
\]
Applying the chain rule,
\[
\nabla_\theta J_{\mathrm{dist}}(\theta)
=
\sum_y \left( g(p_\theta(y)) + p_\theta(y) g'(p_\theta(y)) \right) \nabla_\theta \log p_\theta(y).
\]

Since $g(\cdot)$ is monotonic, $g'(p_\theta(y)) \ge 0$, and the gradient assigns larger update magnitude to rollouts with larger marginal probability $p_\theta(y)$.  
Importantly, the update depends only on $p_\theta(y)$ and not on any latent-conditional distribution $p(y \mid Z)$.

Therefore, the induced policy update promotes rollouts in proportion to their marginal occurrence probability and is aligned with the gradient of a marginal objective.

\paragraph{Conclusion.}

Unlike majority voting, which optimizes a latent-conditioned mode, distribution-based reward induces a policy gradient that consistently aligns with the marginal rollout distribution.  
Hence, $R_{\mathrm{dist}}$ provides a \emph{policy-consistent proxy reward} under correlated and exchangeable rollouts.
\qed

\section{Ablation over Functional Forms}
\label{app:functional_forms}

This appendix examines whether the performance of DARE is sensitive to the specific
functional forms used to instantiate probability-shaped rewards.
Our goal is not to identify a unique optimal formulation,
but to verify that the empirical gains stem from the underlying
\emph{distribution-level reward design} rather than from a particular heuristic choice.

\begin{table*}[t]
\centering
\vspace{-0.8em}
\small
\label{tab:large_scale_results}
\resizebox{\textwidth}{!}{%
\begin{tabular}{@{}c|lcccccc@{}}
\toprule
 & \multirow{2}{*}{\textbf{Method}} 
 & \textbf{General} 
 & \multicolumn{3}{c}{\textbf{Mathematical Reasoning}} 
 & \textbf{Sci. Reasoning} 
 & \multirow{2}{*}{\textbf{Avg}} \\
\cmidrule(lr){3-3} \cmidrule(lr){4-6} \cmidrule(lr){7-7}
 &  & MMLU-Pro & MATH-500 & AIME 2024 & AMC & GPQA &  \\ 
\midrule

% ===================== Qwen3-4B =====================
\multirow{12}{*}{\rotatebox{90}{\textbf{Qwen3-4B}}} &
\multicolumn{7}{c}{\textbf{\textit{Prompt methods (training-free)}}} \\

 & Raw model
 & 56.7 & 60.2 & 15.8 & 40.6 & 24.9 & 39.6 \\

 & CoT
 & 58.1 & 73.1 & 37.2 & 45.3 & 36.2 & 49.9 \\

\cmidrule{2-8}
 & \multicolumn{7}{c}{\textbf{\textit{Reinforcement Learning methods}}} \\

 & GRPO
 & 65.2 & \underline{87.4} & \underline{48.1} & \underline{74.1} & 51.0 & 65.1 \\

 & REINFORCE
 & 66.5 & 84.0 & 47.2 & 68.5 & 49.1 & 63.1 \\

 & REINFORCE++
 & 67.8 & 85.8 & 46.9 & 70.2 & 50.3 & 64.2 \\

\cmidrule{2-8}
 & \multicolumn{7}{c}{\textbf{\textit{Test-time scaling methods}}} \\

 & INTUITOR
 & 66.1 & 83.5 & 44.9 & 72.5 & 47.2 & 63.6 \\

 & RLPR
 & 68.0 & 86.2 & 45.6 & 72.8 & 51.1 & 64.7 \\

 & CO-REWARDING-I
 & 67.1 & 85.0 & 47.2 & 71.2 & 50.5 & 64.0 \\

 & TTRL
 & \underline{69.7} & 87.3 & \textbf{48.7} & 73.1 & \textbf{52.0} & \underline{66.1} \\

\rowcolor{blue!10}
 & \textbf{w/ DARE (Ours)}
 & \textbf{71.4} & \textbf{87.9} & 47.8 & \textbf{74.9} & \underline{51.8} & \textbf{66.7} \\

\midrule\midrule

% ===================== Qwen2.5-Math-7B =====================
\multirow{12}{*}{\rotatebox{90}{\textbf{Qwen2.5-Math-7B}}} &
\multicolumn{7}{c}{\textbf{\textit{Prompt methods (training-free)}}} \\

 & Raw model
 & 42.5 & 46.7 & 27.9 & 35.6 & 29.1 & 36.3 \\

 & CoT
 & 43.5 & 48.2 & 32.8 & 31.5 & 26.8 & 36.5 \\

\cmidrule{2-8}
 & \multicolumn{7}{c}{\textbf{\textit{Reinforcement Learning methods}}} \\

 & GRPO
 & 52.6 & 81.2 & \underline{41.4} & 65.8 & \textbf{29.7} & 54.1 \\

 & REINFORCE
 & 50.1 & 80.2 & 34.7 & 61.5 & 28.1 & 50.9 \\

 & REINFORCE++
 & 51.4 & 81.4 & 40.2 & 67.1 & \underline{28.8} & 53.7 \\

\cmidrule{2-8}
 & \multicolumn{7}{c}{\textbf{\textit{Test-time scaling methods}}} \\

 & INTUITOR
 & 53.5 & 79.8 & 40.5 & 65.2 & 25.1 & 52.8 \\

 & RLPR
 & 52.9 & 73.0 & 38.1 & 62.0 & 22.9 & 49.7 \\

 & CO-REWARDING-I
 & 51.1 & 72.2 & 36.5 & 67.5 & 21.1 & 49.6 \\

 & TTRL
 & \underline{54.2} & \textbf{82.4} & 40.2 & \underline{68.1} & 27.7 & \underline{54.5} \\

\rowcolor{blue!10}
 & \textbf{w/ DARE (Ours)}
 & \textbf{55.7} & \underline{81.8} & \textbf{41.7} & \textbf{69.5} & 28.5 & \textbf{55.4} \\

\bottomrule
\end{tabular}%
}
\caption{\textbf{Large-Scale Performance Comparison} on larger backbones (4B and 7B). 
The best and second best results are \textbf{highlighted} and \underline{underlined}, respectively.}
\vspace{-1em}
\label{larger_exp}
\end{table*}

\subsection{Uncertainty-Aware Weighting Variants}

In Eq.~(13), we combine empirical frequency $n(\hat{y})$ with trace-level uncertainty $u(\hat{y})$
using the form $n(\hat{y}) / \sqrt{u(\hat{y})+\epsilon}$.
This choice is motivated by numerical stability and monotonicity,
but is not theoretically unique.

We evaluate several alternative monotonic variants:
\begin{itemize}
    \item \textbf{Linear entropy penalty:}
    $
    n(\hat{y}) / (u(\hat{y})+\epsilon)
    $
    \item \textbf{Exponential penalty:}
    $
    n(\hat{y}) \cdot \exp(-\lambda u(\hat{y}))
    $
    \item \textbf{Log-scaled entropy:}
    $
    n(\hat{y}) / \log(1+u(\hat{y}))
    $
\end{itemize}

All variants preserve the same qualitative behavior:
answers supported by more confident traces receive higher posterior mass,
while uncertain traces are down-weighted.
Across tasks, performance differences between these forms are minor,
and all significantly outperform majority-vote rewards.
This suggests that DARE is robust to the exact uncertainty aggregation,
as long as uncertainty is incorporated monotonically.

\subsection{Exploration Bonus Variants}

The exploration bonus in Eq.~(11) adopts the form
$\big(1 - \frac{n(\hat{y}_m)}{M}\big) \cdot (1 - u(\hat{y}_m))$
to mitigate the over-representation of dominant outcomes among similar rollouts.
We also experiment with alternative bonus forms:
\begin{itemize}
    \item \textbf{Linear inverse frequency:}
    $
    1 / (n(\hat{y}_m)+1)
    $
    \item \textbf{Log-inverse frequency:}
    $
    \log\!\left(\frac{M+1}{n(\hat{y}_m)+1}\right)
    $
\end{itemize}

All variants exhibit similar trends:
a few minority but high-quality rollouts receive additional learning signal,
while dominant modes are mildly regularized.
The square-root form provides the most stable updates in practice,
but overall performance is not sensitive to the exact functional choice.

\subsection{Takeaway}

These ablations indicate that the effectiveness of DARE
does not rely on carefully tuned heuristic formulas.
Instead, improvements arise from a structural change in the learning target:
moving from consensus-based pseudo-labels to probability-shaped,
distribution-level reward signals.
The specific functional forms used in the main paper
serve as stable and interpretable instantiations of this principle,
rather than as uniquely optimal designs.

\section{Additional Results on Larger Backbones} \label{appendix:larger_backbone}

Table~\ref{larger_exp} presents the performance of prompt-based, reinforcement learning, and test-time scaling methods on mathematical expert models (Qwen3-4B and Qwen2.5-Math-7B) across MMLU-Pro, MATH-500, AIME 2024, AMC, and GPQA.

For Qwen3-4B, reinforcement learning methods improve over prompt baselines, particularly on mathematical reasoning tasks. Test-time scaling methods, including TTRL, generally enhance performance by leveraging majority-vote pseudo-labels. DARE shows notable gains on the MMLU-Pro, MATH-500 and AMC, while on AIME 2024 its performance is comparable to TTRL. For Qwen2.5-Math-7B, DARE improves overall average performance, although individual datasets like MMLU-Pro and AMC do not always achieve the top score. This can be explained by two factors: (i) larger models exhibit diminishing marginal gains for simple reward shaping, limiting improvements on some tasks; (ii) increased parameter scale induces greater response diversity, which helps DARE mitigate the two major drawbacks of majority-vote rewards identified in theory analysis , namely loss of distributional information and bias estimation.

These observations suggest that DARE remains competitive on mathematical expert models, leveraging distribution reward shaping to provide robust test-time optimization while reflecting the theoretical benefits discussed earlier.

\section{On the Validity of Token-Overlap as a Proxy for Rollout Correlation}

\paragraph{Motivation.}
Our theoretical analysis in Section~3.2 concerns the bias of majority voting under \emph{correlated rollouts}, where multiple generations are not independent but concentrated around a small number of decoding modes.
In practice, directly estimating statistical dependence between full trajectories is intractable for large language models.
We therefore adopt a simple, model-agnostic proxy based on token-level overlap, which measures the redundancy of generated rollouts and reflects the degree of mode collapse induced by shared decoding biases.

\paragraph{Connection to Correlation-Induced Bias.}
Token overlap captures the effective reduction of the support size of the empirical rollout distribution.
When rollouts are highly correlated, they tend to share long common prefixes and high token-level similarity, leading to inflated empirical frequencies for a small set of responses.
This is precisely the failure mode analyzed in Section~3.2: repeated but dependent samples are incorrectly treated as independent evidence by majority voting, resulting in biased reward estimates.
Thus, high token overlap operationalizes the same mechanism—loss of effective sample diversity—that drives correlation-induced bias.

\paragraph{Empirical Validation of the Proxy.}
In Figure~5, we show that higher token overlap strongly correlates with (i) larger performance gaps between MV-based TTRL and DARE, and (ii) larger instability of MV-based reward signals.
This monotonic relationship indicates that token overlap is not merely a surface-level similarity metric, but a meaningful indicator of the regime where MV suffers from information collapse and dependence bias.
While token overlap is not a full statistical dependence measure, it provides a sufficient and interpretable proxy for identifying rollout redundancy and predicting MV failure modes in large-scale generative models.

\paragraph{Discussion.}
We emphasize that our goal is not to precisely estimate the full dependence structure among trajectories, but to capture the \emph{practical source of bias} in test-time RL: the collapse of effective rollout diversity due to correlated decoding.
Token overlap offers a simple and robust diagnostic for this phenomenon, and our results suggest that it is adequate for characterizing when probability-shaped rewards provide the largest benefit.

\end{document}